Accepted at IEEE Instrumentation & Measurement Magazine (2026)

# Lost in Motion: Vision Language Models Fail the Dynamic Gauges Test

Tairan Fu, Francisco Javier Santos-Martín, Javier Conde, Elena Merino-Gómez, Pedro Reviriego

The digital transformation of industrial manufacturing increasingly relies on the ability of autonomous robots to interact with legacy infrastructure, particularly analog gauges. Vision-Language Models (VLMs) have the potential to provide a general solution for gauge reading and have already shown good performance in instrument recognition. However, performing accurate, real-time gauge readings is a more complex task. This paper evaluates state-of-the-art models, including versions from the GPT-5 (5.4 Thinking and 5.3 Instant) and Gemini 3 (Pro and Flash) families, against a set of simple but realistic dynamic gauge reading scenarios. To facilitate this evaluation, we introduce a novel dataset comprising video sequences of three instruments of different gauge types: circular, linear, and Vernier, under diverse motion and speed profiles. Our findings indicate that the evaluated frontier VLMs, under our specific testing conditions, exhibit a limited ability to interpret needle trajectories and scale semantics, failing to provide the traceability and reliability needed for safety-critical monitoring. The results demonstrate that these models have not yet achieved the performance necessary to be classified as trustworthy synthetic instruments under existing IEEE and ISO standards.

## Introduction

Artificial Intelligence (AI) is becoming a fundamental component of modern instrumentation, shifting the paradigm from simple data processing to AI-defined measurement systems. This is particularly evident in industrial manufacturing, where autonomous robots are increasingly tasked with high-precision monitoring in complex environments [1]. The adoption of machine learning and AI for Instrumentation and Measurement (I&M) has allowed for the automation of tasks that were previously reliant on human perception, such as defect detection and instrument reading. However, as these algorithms become as critical to the measurement result as the sensor hardware itself, the industry faces new challenges in ensuring that these synthetic instruments maintain the rigorous standards of reliability, traceability, and uncertainty quantification required in metrology. This challenge is particularly evident in industrial manufacturing, where autonomous robots must interact with legacy infrastructure such as analog gauges [2]. Traditional approaches to read gauges typically utilize edge detection and the Hough Transform to isolate gauge dials and their positions providing high frequency sampling and a deep understanding of the algorithms used which enables the analysis of errors [3],[4]. However, since they tend to use heavily customized pipelines for a given environment they struggle in non-ideal "open field" environments.
The emergence of Vision-Language Models (VLMs) capable of analyzing complex video sequences positions them as natural candidates to be used as virtual instruments. The question is whether VLMs will be capable of translating visual data from analog gauges, such as circular, linear, and Vernier type, into digital readings [5]. VLMs introduce sophisticated probabilistic reasoning and could in principle deal with general scenarios; however, they are a black box whose performance is hard to analyze and understand. The integration of VLMs into Instrumentation and Measurement raises critical questions regarding reliability, traceability, and standardization. While state-of-the-art models demonstrate impressive zero-shot capabilities in static instrument recognition, they may fail in more realistic settings that are not static [6]. In dynamic environments, where continuous operational transients and rapid needle movements are common, these models must function not just as vision systems, but as precise measurement instruments. Dynamic measurement requires models to capture continuous frame-to-frame changes, introducing a temporal resolution challenge. If the model fails to keep pace with this physical motion, temporal aliasing occurs. Furthermore, the models also need to correctly interpret not only each frame but also put them in context with the previous ones while measuring the time difference, which is a significantly more complex task than analyzing a static image. Additionally, to be valid, an AI-based measurement system must provide reliable readings under different conditions and for different gauge types.
To answer those questions, this article explores the potential of state-of-the-art Video-Language Models for the task of Dynamic Analog Gauge Reading, moving beyond static snapshots to investigate how models interpret temporal change and motion dynamics. We emphasize that the findings and metrological conclusions presented in this study are strictly bounded to the evaluated model versions and the specific dynamic configurations of our experimental setup.

**Vision Language Models in Instrumentation and Measurement**

VLMs are pre-trained on massive datasets combining textual descriptions, static images, and continuous video streams. This enables them to perform sophisticated reasoning so, they can identify, interpret, and extract semantic data from visual environments they have never explicitly encountered during training, entirely guided by high-level natural language instructions known as prompts [7]. For Instrumentation and Measurement, VLMs represent an appealing candidate to serve as flexible, out-of-the-box virtual instruments. In principle, a single foundational model could be deployed across an entire factory floor, seamlessly adapting to circular pressure dials, linear fluid columns, and high-precision Vernier calipers without requiring custom code or geometry-specific calibration pipelines. This flexibility would represent a major leap forward for autonomous robotic inspection, allowing mobile platforms to interact with legacy analog hardware much like a human operator would.
However, substituting deterministic vision pipelines with foundational VLMs introduces a profound paradigm shift in measurement science. Classical instrumentation relies on a clear, unbroken chain of traceability where errors are mathematically bounded and physically traceable to established standards. VLMs, by contrast, are inherently probabilistic black-box systems. They generate data outputs based on statistical correlations and multi-layered neural networks, rather than explicit physical geometry. Consequently, their decision boundaries are non-linear, difficult to audit and even harder to explain. For the instrumentation community, this creates a conflicting scenario: while VLMs potentially possess the flexible spatial reasoning needed to navigate complex, real-world open-field environments, their nature makes it challenging to guarantee the strict reliability required for safety-critical monitoring. To consider these models as valid virtual instruments, they must meet fundamental metrological requirements. This demands establishing clear pathways for traceability, which are currently obscured by the black-box nature of the models. It also requires robust frameworks for uncertainty quantification, as the probabilistic nature of VLMs prevents the straightforward mathematical error bounding typical of physical sensors. Finally, ensuring the validation and trustworthiness of such systems necessitates standardized, dynamic evaluation protocols that can guarantee predictable behavior under operational transients.
To move forward, the instrumentation community must be able to empirically measure VLM performance under realistic, time-series operational conditions. However, conducting a rigorous metrological evaluation is currently impossible because there are no publicly available video datasets that combine dynamic analog gauge configurations with high-fidelity and grounded temporal references. To overcome this bottleneck and establish a standardized benchmark for multimodal AI systems, we developed the Dynamic Gauge Dataset (DGD). The following section summarizes the architecture, mechanical motion profiles, and data quality design principles of this reference repository.

**The Dynamic Gauge Dataset (DGD): A Reference for VLM Dynamic Gauge Reading**
Traditional benchmarks focus on the evaluation of static images [3], [7] and do not account for the complexity of temporal variations which are key in real applications with small and fast vibrations that are not easy to capture in images. To the best of the authors' knowledge only one video dataset is publicly available which has only ten videos for a single instrument type, with no in-band time reference and no detailed metadata [4]. Therefore, for our experiments we collected a dataset of videos specifically designed to assess the temporal tracking capabilities of Vision-Language Models. The result is the Dynamic Gauge Dataset (DGD) that covers three analog instruments that represent a spectrum of metrological complexity gauges:

1) **Circular Dial**: a standard rotating pointer interface where visual interpretation requires precise angular mapping and parallax compensation.
2) **Linear Scale**: a one-dimensional sliding indicator representing basic translational displacement.
3) **Vernier Scale**: a high-precision instrument that demands sub-pixel alignment logic across dual sliding scales, providing a stress-test for the model's geometric resolution.

From a metrological perspective, the dataset prioritizes temporal grounding and traceability. Gauge motion is not arbitrary but is actuated at predefined mechanical speeds to ensure reproducible dynamics. Each sequence is recorded at 30 frames per second (fps) under stabilized illumination to minimize uncertainty from motion blur or flickering. For the movement of the instruments, a universal electromechanical testing machine, brand: SHIMADZU, model: AG-100KN-MS, controlled with the Trapezium 2 software, was used. An important feature is the inclusion of an in-band digital chronometer overlaid on the video stream. This chronometer serves as a temporal reference standard that ensures traceability and accuracy, two core dimensions of the ISO/IEC 25024 standard requirements for data quality [8]. By embedding precise timestamps directly into the video, we eliminate uncertainties stemming from variable frame rates providing an immutable "reference clock" for the measurement process. To ensure that tracking failures were not caused by a failure to interpret the temporal reference, we independently verified that all evaluated models could read the in-band digital chronometer with near-perfect accuracy.

Figure 1 illustrates the experimental setup, showcasing the three instrument specimens and the integrated in-band digital chronometer. Each video in the DGD is accompanied by a standardized JSON metadata file that serves as its metrological "digital twin". This structured schema, also aligned with the data quality dimensions of ISO/IEC 25024, decouples visual content from physical parameters such as gauge type, motion speed, and movement direction. The metadata includes the values of the gauge reading every 200 ms. The values were obtained using 10 keyframes per video sequence that were manually and visually annotated by human operators, with the intermediate frame values computed via linear interpolation across these fixed references. From a metrological standpoint, we acknowledge that manual annotation and subsequent linear interpolation introduce an implicit baseline uncertainty that is discussed and quantified in a supplemental document [9] to be below 0.5% of the measurement range. As demonstrated during testing, the massive structural errors exhibited by the evaluated VLMs, mean that uncertainty has a negligible impact on the quantitative and qualitative results. The videos and metadata are publicly available[1] to facilitate further research and VLM evaluation as well as the extension of the dataset to cover a wider range of gauge, dynamic and environmental configurations.

In its initial version, the dataset contains 54 videos, eighteen per gauge type. The motion profiles are specifically designed to provide an initial step for benchmarking Video-Language Models. Three simple patterns are used: increasing monotonic, decreasing monotonic, and first increase, then decrease. In the first two patterns, the gauge indicators move in a single, consistent direction throughout the duration of the clip, while in the third they move in the both directions adding additional complexity to the reading [9]. By focusing on simple movements, the experiment eliminates the noise of unpredictable movements and back and forth oscillations. This simplicity is intentional. It allows researchers to isolate a model's ability to synchronize visual motion with time without the added complexity of tracking fast direction changes. The only direction change occurs in the third pattern but is a single change in the middle of the video not an oscillation. By establishing this baseline, it becomes much easier to quantify "temporal blindness" and determine if a model can accurately map a steady physical change to the passing seconds on the in-band digital chronometer.

The dataset covers a wide spectrum of these three profiles by utilizing different speeds. For example, in instruments equipped with a Vernier, the displacement speeds range from a slow movement of 5 divisions per second to a fast speed of 100 divisions per second, which implies a speed interval ranging from 3 to 60 mm/min for the analog dial indicator (resolution of 0.01 mm) and from 15 to 300 mm/min for the analog depth gauge (resolution of 0.05 mm). In the case of the depth gauge without a Vernier (resolution of 1 mm), the displacement speeds range from 0.5 to 10 scale divisions per second, which translates into a speed interval between 30 and 600 mm/min. These variations test the limits of how well a VLM can maintain precision as the rate of visual change increases. These foundational videos are intended to serve as the starting point and are described in more detail in [9]. Once a model demonstrates that it can reliably perform on these straightforward patterns, the benchmark can be evolved. More complex patterns, such as needle vibrations, sudden stops, or fast reversals, can then be introduced to further challenge the model's temporal reasoning and physical understanding.

[1] The dataset is available at: https://zenodo.org/records/20624895.

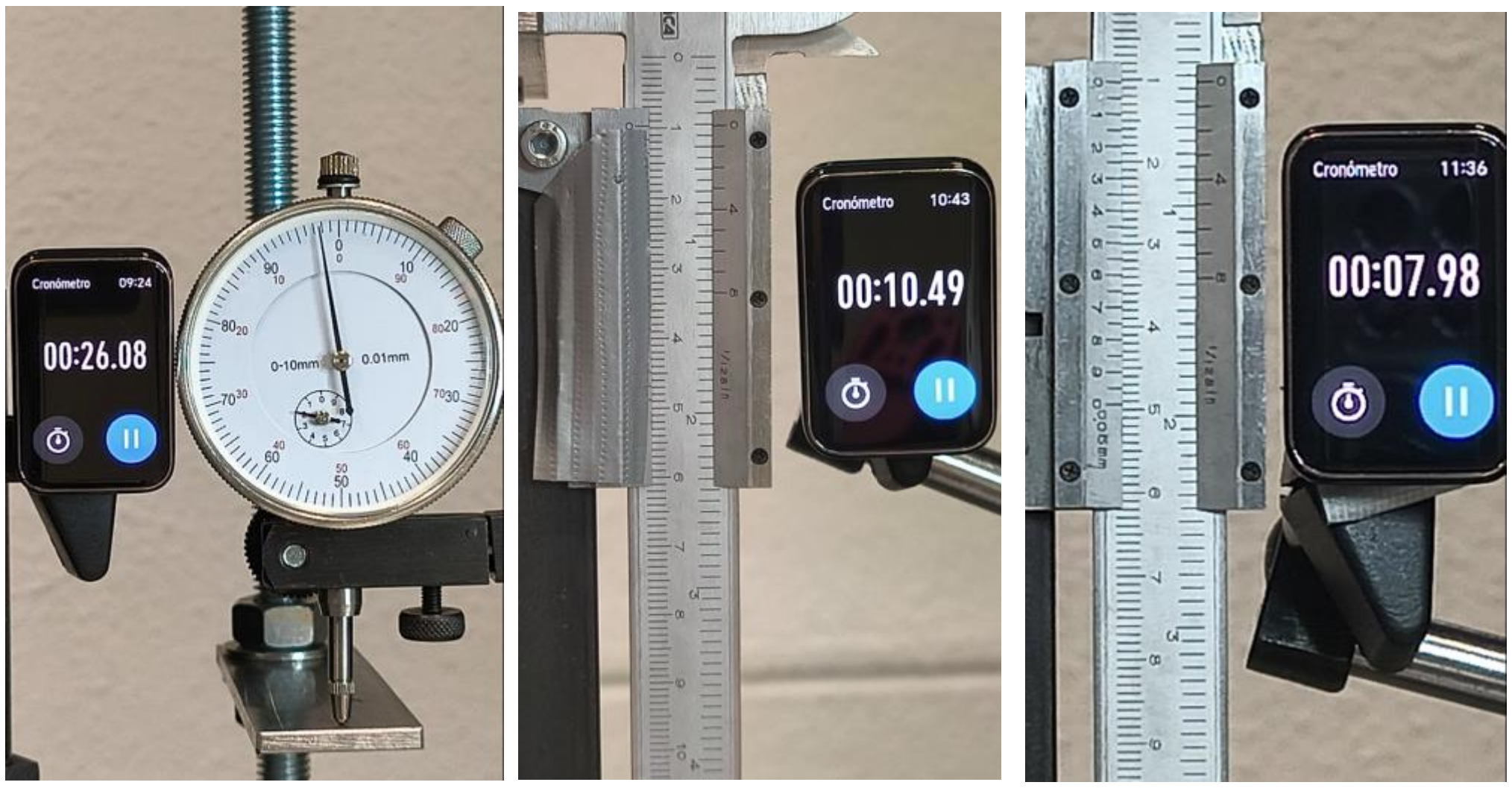

***Fig. 1.*** Illustrations of the three instruments of different gauge types in the DGD dataset. From left to right: (a) uniform angular rotation, (b) linear motion, (c) Vernier. The video snapshots show the in-band digital chronometer.

### *Can Frontier Models Read Dynamic Gauges?: Testing VLMs on the Dynamic Gauge Dataset*

To put Vision-Language Models (VLMs) to the test, we selected a set of frontier models that includes reasoning systems and native multimodal architectures. These models represent the state-of-the-art of VLMs at the time of writing this paper on an attempt to achieve the best possible performance. The models selected[2] are: GPT-5.4 Thinking and GPT-5.3 Instant from OpenAI and Gemini 3 Pro and Gemini 3 Flash from Google. The models were configured to control variability in an effort to ensure similar results every time the test is run[3]. Fine-tuning the models was out of scope, as the primary objective was to evaluate off-the-shelf models and is left for future work. The code and model settings used in the experiments are publicly available [4] to ensure reproducibility.

A structured prompting strategy based on the Visual Chain-of-Thought (VCoT) paradigm was used to break down the complex task into three intuitive stages following an intuitive 'See-Think-Confirm' logic [10]. This approach uses three consecutive stages: See, where the model calibrates visually by identifying gauge edges and scale markers to avoid confusion from reflections; Think, where it compares sequential frames to track needle motion over time rather than guessing from a single snapshot; and Confirm, which serves as a final reality check to verify that the readings remain physically consistent between frames. Additional details of the prompts and experimental procedure are provided in the supplemental material document.

The experimental results reveal that performance was poor across the board, with models generally struggling to maintain temporal grounding and geometric rigor. In most cases, the models failed to follow the basic directional trend of the moving needle, frequently exhibiting massive offsets or incorrect slopes compared to the ground truth. The exception was GPT-5.4 Thinking, which achieved high precision in specific configurations. However, such instances were rare, rendering the evaluated frontier models currently unreliable under these specific testing conditions for formal industrial metrology. To illustrate the behaviour of the models, we selected a few representative examples that are shown in Figures 2, 3, and 4. Figure 2 shows an example in which GPT-5.4 Thinking is able to track the gauge while the other three models fail in different ways, despite all being able to correctly read the gauge at the start of the test before the value begins to increase. The failures manifest as constant readings, temporarily increasing readings that abruptly drop back to the initial baseline value after a few frames, or a correct directional trend but with an incorrect slope.

[2] The exact names of the models are: gpt-5.4-2026-03-05, gpt-5.3-chat-latest, gemini-3-flash-preview, gemini-3-pro-preview. Note that even if gpt-5.3-chat-latest, gemini-3-flash-preview and gemini-3-pro-preview do not include static date-coded suffixes, they represent official API releases that uniquely identify the specific model.

[3] To quantify run-to-run variance, one configuration was executed ten times using Gemini 3 Flash. The normalized MAE was 14.15% ± 3.54% and the Pearson correlation was 0.606 ± 0.11, yielding an identical fail verdict across all ten runs and confirming that variability is small relative to the failure margin.

[4] The code and evaluation scripts including model settings are available at: https://github.com/aMa2stai210/VLM_gauge

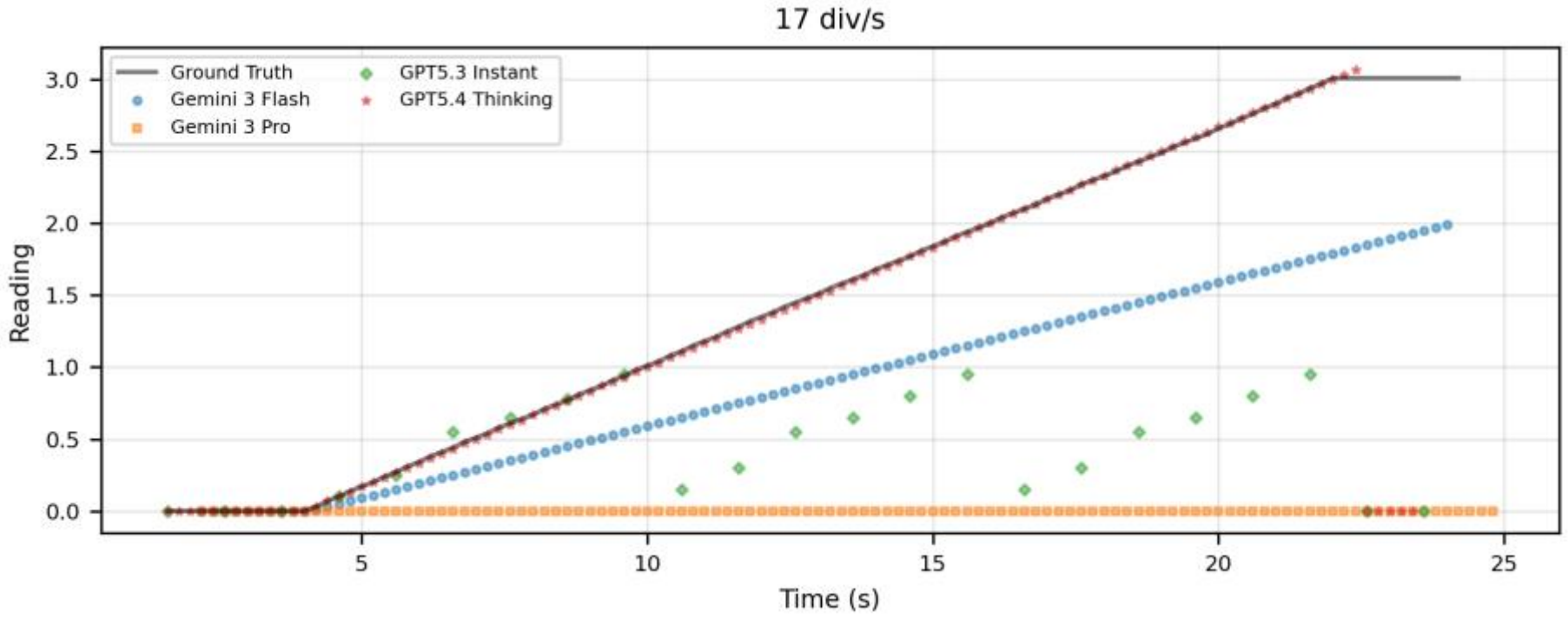

***Fig. 2.*** Example of results for angular rotation illustrating the ability of GPT-5.4 Thinking to track the gauge while the other three models fail to detect the movement (Gemini 3 Pro), producing erratic readings (GPT-5.3 Instant) or following the movement but with a lower speed (Gemini 3 Flash)

The results in Figure 3 for a linear gauge and the increase/decrease motion pattern illustrate a scenario in which all models fail to produce meaningful readings. A common limitation of the models is their inability to even synchronize on the initial baseline value of the gauge. This initial failure sets an incorrect starting point, which completely skews the subsequent time-series data from the very first frame. Furthermore, even when a model manages to capture the starting value, it often completely fails to follow the physical trend of the needle. Another common pattern is the response in steps that with several patterns in the readings creating staircase like plots when each of the patterns is a constant value. It is worth noting that the steps in the staircase last for several consecutive samples, even though the physical gauge pointer moves monotonically, taking a distinct value at each sample. Therefore, this pattern cannot be attributed to the 5 Hz frame sampling used to feed the models, but instead represents an intrinsic limitation of the models, which possess the visual information necessary to detect the frame-by-frame gauge movement yet fail to update their temporal state.

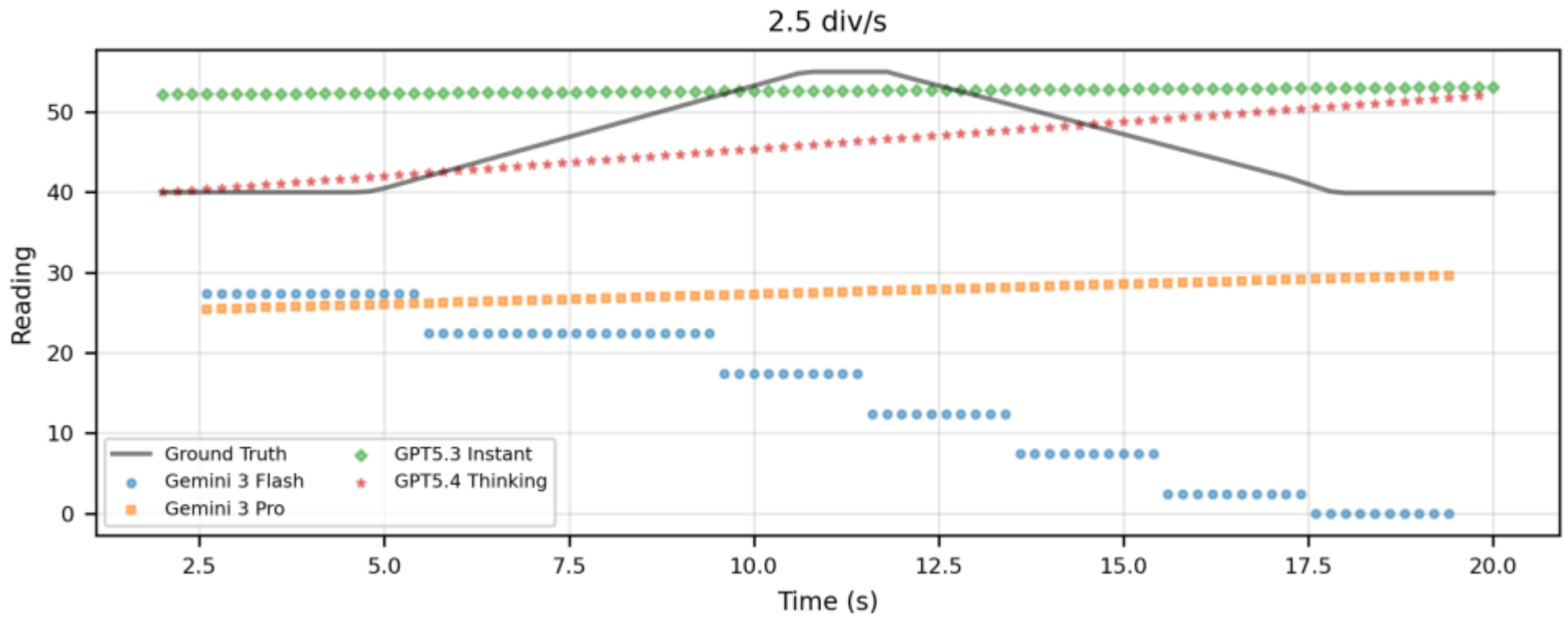

***Fig. 3.*** Example of results for the linear scale gauge illustrating the inability of all the models to produce meaningful readings. GPT-5.4 Thinking and Gemini 3 Pro have continuously increasing readings, Gemini 3 Flash a descending /staircase pattern and GPT-5.3 Instant an almost constant reading.

Finally, Figure 4 presents the results for a Vernier gauge showing again the inability of the models to track the readings for a descending movement pattern and exhibiting similar failure modes like constant readings or staircase patterns.

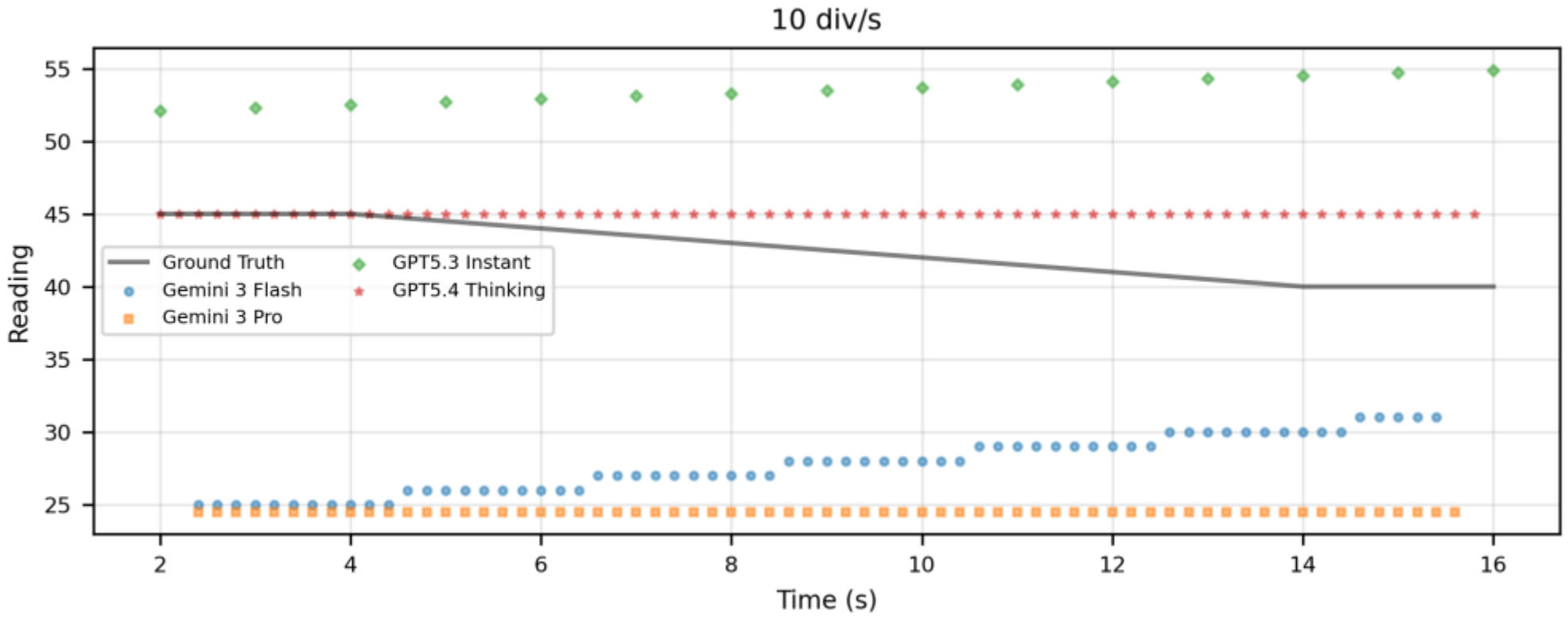

***Fig. 4.*** Example of results for the Vernier scale gauge illustrating the inability of all the models to produce meaningful readings. GPT-5.4 Thinking and GPT-5.3 Instant have continuously increasing readings, Gemini 3 Flash a increasing staircase pattern and Gemini 3 Pro an almost constant reading.

The results for all the configurations are available in the supplemental document [9] and as the examples in Figures 2,3,4 show several recurrent patterns that may explain some of the limitations of the models: horizontal flat-lines and staircase jumps. In the first case, the model output remains unchanged (a horizontal line) even while the ground truth is moving. In the second, horizontal segments are followed by jumps, forming a staircase shape. This suggests that the models do not process the video as a continuous stream but instead extract a few snapshots to process. The flat line likely occurs because the model lacks visual data for the time between these snapshots or fails to see the change as it is too small. Then it may simply repeat the last value, effectively blinking through the motion. The staircase jump likely happens only when the next sampled snapshot is processed, and the model finally detects that the needle has moved. These patterns suggest that the failure might not just be due to the speed of the needle, but rather a lack of temporal continuity. If the model treats a dynamic measurement as a series of disconnected still images, it cannot provide reliable tracking. The experiments, by identifying these artifacts, provide a diagnostic roadmap for improving how future models integrate motion.

To quantitatively evaluate performance, we established a pass/fail threshold of 5% normalized Mean Absolute Error and a 0.95 Pearson correlation coefficient. Under these criteria, all evaluated models failed across all configurations, with the sole exception of GPT-5.4, which successfully passed in the opening and closing settings of the linear depth gauge. These results demonstrate that while frontier VLMs can qualitatively describe a gauge, they lack the temporal grounding and geometric rigor required for certified metrology[5] under the tested dynamic scenarios. The DGD test proves that massive parameter counts do not automatically translate into the millisecond-level precision needed for industrial control loops.

[5] Additional details including the normalized mean absolute error and the standard deviation are included in the supplemental material document.

### *Closing the Loop: Limitations and Future Directions in Dynamic Gauge Reading*

Our experimental results show that the evaluated frontier models struggle with dynamic gauge reading under our testing conditions. While they excel at qualitative scene description and other tasks, they struggle to maintain the traceable precision required for industrial instrumentation. In this context, the Dynamic Gauge Dataset (DGD) can be used as a test and a standardized metric for measuring the progress of future VLMs. By providing a controlled environment with millisecond-level ground truth, the DGD allows researchers to quantify progress, enabling the evaluation of future models for dynamic gauge reading.

This work is only the initial step towards VLM-based dynamic gauge reading and has a number of limitations that include:

1) **Idealized Conditions**: the current dataset features optimal viewing angles and lighting to isolate reasoning capabilities, which does not fully represent the complex and non-ideal reality of factory floors.
2) **Sampling Frequency**: our 5 Hz extraction protocol, while sufficient for many tasks, may not be sufficient for ultra-high-speed transients.
3) **Dataset Scale and Diversity**: the current version is constrained in scale, featuring a specific number of videos per instrument category and on a single instance of each instrument. Expanding the volume of data is necessary to ensure statistical robustness across the diverse edge cases found in global manufacturing.
4) **Instrument Type:** the study focuses on three primary gauge morphologies (Circular, Linear, and Vernier). However, industrial environments utilize a vast array of specialized indicators, such as multi-pointer pressure gauges, drum-type counters, and liquid-level sight glasses, which are not yet represented in the DGD.
5) **Ground-Truth and Annotation Uncertainty:** the reference metadata relies on human manual keyframe annotation and subsequent linear interpolation, which introduces a baseline uncertainty. While this is acceptable for exposing the catastrophic tracking failures of current VLMs, it inherently limits the dataset's capacity to evaluate fine-grained visual tracking alignments.

To progress towards reliable VLM-based dynamic gauge reading future research should consider:

1) **Expansion of the DGD:** increasing the dataset's scale beyond the current number of videos per category to ensure statistical robustness both in speed and gauge instance.
2) **Gauge Taxonomy:** Incorporating specialized industrial indicators such as multi-pointer gauges and drum-type counters.
3) **Environmental Robustness:** integrating vibration, motion blur, and extreme glare to move beyond idealized conditions.
4) **Internal Temporal Logic**: developing architectures that derive "internal" time-sense from frame-level features without requiring an in-band chronometer.
5) **High-Precision Ground-Truth:** replacing manual keyframe interpolation with continuous tracking systems to reduce reference uncertainty.

While the results from the DGD test with the evaluated models and configurations suggest that we are currently far from achieving reliable, VLM-based dynamic gauge reading, the rapid development of generative artificial intelligence suggests a rapidly closing window. The transition from static image understanding to native video reasoning has occurred with unprecedented speed in a matter of months. Therefore, the gap in temporal grounding and geometric rigor may be closed sooner than our results suggest, and we may soon have models that excel at the DGD test. Finally, based on our empirical findings, we outline several practical lessons and design guidelines for instrumentation engineers:

1) **Avoid Pure End-to-End VLM Pipelines for Real-Time Loops:** Engineers should not rely solely on end-to-end VLMs for safety-critical or high-frequency control loops. The intrinsic "temporal blindness" and staircase tracking artifacts observed in frontier models mean they cannot guarantee the millisecond-level determinism required by industrial standards.
2) **Implement Hybrid Architectures:** Traditional deterministic pipelines (e.g., edge detection and Hough transforms) are highly reliable for localized, high-frequency coordinate tracking and error analysis. VLMs are best reserved for higher-level semantic tasks, such as initial instrument type recognition, context-aware calibration, and anomaly detection.
3) **Enforce Deterministic Validation Wrappers:** Any AI-based virtual instrument must be wrapped in a deterministic, rule-based software layer. This layer should run real-time sanity checks (such as physical monotonicity constraints and maximum rate-of-change limits) to automatically intercept and flag the severe offsets or non-physical tracking jumps characteristic of probabilistic black-box models.

## Conclusion

This paper demonstrates that, under the specific testing conditions evaluated, the selected frontier VLMs exhibit significant tracking limitations and are currently not capable of reliable dynamic gauge reading. Results from the Dynamic Gauge Dataset (DGD) created as part of this work reveal a significant precision gap where models struggle with temporal grounding, physical monotonicity, and high-speed needle tracking. The study's primary limitations include the use of idealized conditions with optimal lighting and angles that do not reflect the noisy reality of factory floors. The instrument diversity is also currently restricted to Circular, Linear, and Vernier scales, excluding specialized industrial indicators like multi-pointer gauges. Future research should focus on expanding the DGD scale, integrating environmental noise such as vibration and glare, and developing models with internal temporal logic that do not rely on visual chronometers. While the models evaluated are not yet reliable for dynamic reading, the rapid evolution of generative AI suggests this gap may close sooner than expected. The transition from static to native video reasoning has occurred in a few months, indicating that models capable of excelling at the DGD test may arrive in the near future.

## Acknowledgment

This work was supported by the Agencia Estatal de Investigación (AEI) (doi:10.13039/501100011033) under Grants FUN4DATE (PID2022-136684OB-C22) and SMARTY (PCI2024-153434), by TUCAN6-CM (TEC-2024/COM460) funded by CM (ORDEN 5696/2024) and by the BRIDGE-AI (Grant 101299050) project funded by the European Union's Horizon Europe research and innovation program. The access to the models was provided by the OpenAI researcher access program and from Google.org, and the Google Cloud Research Credits program for the Gemini Academic Program.

***Tairan Fu*** (tairan.fu@polimi.it) is a PhD student at Politecnico di Milano working on the use of AI in manufacturing.

***Francisco Javier Santos-Martín*** (francisco.santos@uva.es) is an associate professor at the Universidad de Valladolid. His research focuses on manufacturing engineering, metrology, and standardization.

***Javier Conde*** (javier.conde.diaz@upm.es) is an assistant professor at the Universidad Politécnica de Madrid and a researcher at the Information Processing and Telecommunications Center. His research focuses on digital twins and evaluation of AI models.

***Elena Merino-Gómez*** (elena.merino.gomez@uva.es) is an associate professor at the Universidad de Valladolid. Her research focuses on manufacturing engineering, metrology and AI.

***Pedro Reviriego*** (pedro.reviriego@upm.es) is a professor at the Universidad Politécnica de Madrid and a researcher at the Information Processing and Telecommunications Center. His research focuses on the design and evaluation of AI models and systems with a focus on reliability and security.